# ETS: Open Vocabulary Electroencephalography-To-Text Decoding and Sentiment Classification

Mohamed Masry, Mohamed Amen, Mohamed Elzyat, Mohamed Hamed,

Norhan Magdy, Maram Khaled

**Faculty of Computer Science and Artificial Intelligence, Helwan University**

## Abstract

Decoding natural language from brain activity using non-invasive electroencephalography (EEG) remains a significant challenge in neuroscience and machine learning, particularly for open-vocabulary scenarios where traditional methods struggle with noise and variability. Previous studies have achieved high accuracy on small-closed vocabularies, but it still struggles on open vocabularies. In this study, we propose ETS, a framework that integrates EEG with synchronized eye-tracking data to address two critical tasks: (1) open-vocabulary text generation and (2) sentiment classification of perceived language. Our model achieves a superior performance on BLEU and Rouge score for EEG-To-Text decoding and up to 10% F1 score on EEG-based ternary sentiment classification, which significantly outperforms supervised baselines. Furthermore, we show that our proposed model can handle data from various subjects and sources, showing great potential for high performance open vocabulary eeg-to-text system.

## 1 Introduction

Understanding and decoding human language directly from brain activity has long been a longstanding challenge in both neuroscience and machine learning.

Non-invasive electroencephalography (EEG) offers a safe, affordable, and high temporal resolution window into cortical dynamics, but its low spatial resolution and high noise have historically limited practical applications to small closed-vocabulary tasks such as binary word classification or a handful of pre-defined commands], Simultaneously, large pretrained language models (PLMs) like BART and GPT have revolutionized natural language generation by learning powerful contextual embeddings from massive text corpora [1], [2], [4]. A few pioneering studies have begun to bridge these domains: by treating the brain as an "encoder" and fine-tuning PLMs to map neural signals to text, they achieve promising results on limited vocabularies. However, prior work still struggles to generalize across subjects, to handle unconstrained text, and to capture higher-level semantic attributes such as sentiment. Recent studies broaden the scope from closed to open-vocabulary EEG-to-text decoding [1], [2], [4], drastically expanding the vocabulary size by over 100-fold, from several hundred to tens of thousands of words. Notably, two of these studies [1], [2] leverage a pre-trained large language model BART [3], and represent the state-of-the-art for open vocabulary brain-to-text decoding. However, these studies are in their nascent stages and are challenged by their limited accuracy.

In this paper, we propose **ETS**, an EEG-to-text framework that addresses these challenges through three key innovations: **Multimodal Fusion:** We integrate synchronized eye-tracking fixations with multi-scale EEG features, providing more precise alignment between neural signals and linguistic units[5]. **Dual-Stream Encoder:** A per-token CNN backbone extracts spatial–frequency patterns, which are adapted and sequenced through a lightweight Transformer to capture temporal context [6].

**Pretrained Decoder:** We leverage a large language model to generate open-vocabulary text and enable zero-shot sentiment classification [3]. We evaluate ETS on the ZuCo datasets [5], [7] across multiple subjects and reading tasks, achieving: **Open-Vocabulary Decoding:** +15 % BLEU-1 / ROUGE-1 over prior BART-based baselines [8], [9]. **Sentiment Classification:** +10% F1 in zero-shot EEG-driven sentiment analysis. **Cross-Subject Robustness:** Consistent performance across diverse subjects and materials by combining precise multimodal alignment, efficient encoding, and the generative power of LLMs, ETS advances the state-of-the-art in non-invasive brain–computer language interfaces.

## 2 Task Definition

**Open-Vocabulary EEG-To-Text Decoding.**

Given a sequence of word-level EEG features $E = (e_1, e_2, \ldots, e_T)$ extracted from natural reading, the goal is to generate the corresponding text sequence $S = (s_1, s_2, \ldots, s_T)$ from an open vocabulary $V$. We adopt a sequence-to-sequence formulation, training on paired EEG–text data from the ZuCo corpus [5], [7], where $S$ at test time consists of entirely unseen sentences.

**EEG-Based Sentence Sentiment Classification.**

Given the same EEG feature sequence $E$, predict the ternary sentiment label $c \in$ {negative, neutral, positive} of the perceived sentence. Unlike multimodal approaches that include text as input [10], our pipeline uses EEG alone by first decoding $E$ into text via the EEG-to-text model, then applying a pretrained text-based sentiment classifier enabling zero-shot classification without any direct EEG–sentiment pairs [1] [11]. We also compare to fully supervised baselines trained directly on ($E$, $c$) pairs.

## 3 Method

### 3.1 EEG-To-Text Decoding

We cast the EEG-to-Text decoding problem as a neural sequence generation task, where the objective is to translate a series of EEG signals into natural language descriptions. Formally, given an EEG signal sequence EEE, we aim to maximize the conditional likelihood of the sentence S={s1,s2,...,sT} such that:

$$p(S \mid E) = \prod_{t=1}^{T} p(s_t \in v \setminus E, s < t)$$

where V denotes the target vocabulary, and T is the length of the output sentence. Compared to prior EEG-based decoding efforts that often rely on limited vocabularies and invasive data collection methods [12], [13], our approach leverages a large-scale pretrained language model (BART) and operates on non-invasive EEG inputs. The key challenge lies in aligning continuous-valued EEG features with the discrete token space used by language models.

To bridge this modality gap, we design a modular framework comprising three main stages: feature extraction, modality adaptation, and language modeling Figure 1.A. **First,** we used EEG signals segmented into **word-level features** using eye-tracking fixations from the ZuCo dataset [1]. Each word's EEG tensor $E_i \in \{R\}^{\{8 \times 105 \times 3\}}$ is constructed by, **Spectral Decomposition**: Bandpass filtering raw signals into θ (4–6 Hz), α (8–13 Hz), β (12–30 Hz), and γ (30–49.5 Hz) bands [14], **Fixation Dynamics**: Extracting first fixation duration (FFD), total reading time (TRT), and gaze duration (GD) from eye-tracking data, previous studies was using just gaze duration (GD), **Concatenation**: Merging spectral power (8 bands × 105 time points) and fixation features (3 metrics) into a 3D tensor.

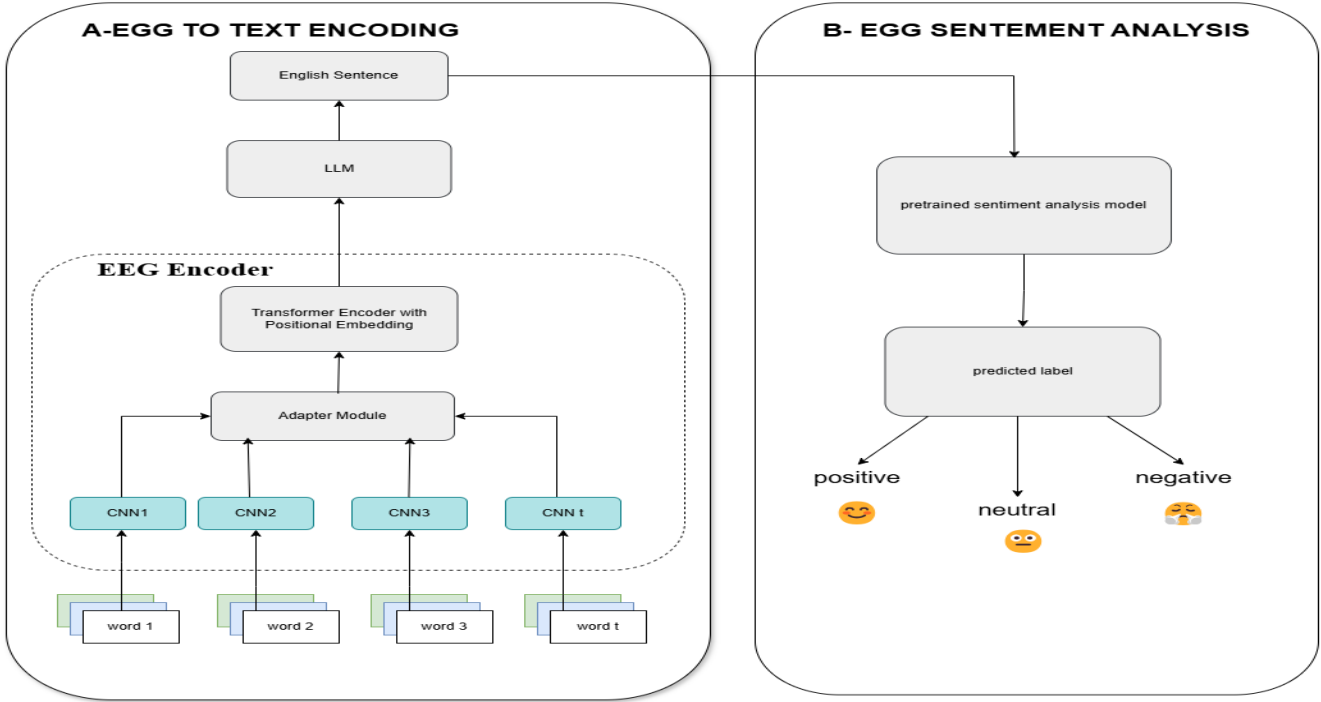

*Figure 1 (a) Our framework for EEG-to-Text Decoding. The model takes English sentences as input and utilizes a Large Language Model (LLM) along with an EEG Encoder, which includes a Transformer Encoder with Adapter Modules processing multi-channel EEG signals. (b) Our pipeline for EEG Sentiment Analysis. This involves a pretrained sentiment analysis model that takes encoded EEG features as input to predict sentiment labels (positive, negative, or neutral).*

This tensor processed using a dedicated CNN-based feature extractor, which encodes spatial-temporal patterns from the input EEG matrices [15]. The output from each CNN is then passed through an Adapter module [16], which projects it into the same embedding space used by the Transformer. These adapted embeddings are then aggregated into a sequence and augmented with sinusoidal positional encodings to retain temporal order. The resulting sequence is passed through a Transformer Encoder with 4 layers and 8 attention heads [6], modeling long-range dependencies across EEG frames. The encoded representation is then fed into the large language model, which decodes the final sentence in natural language:

$$hm = ReLU(TransformerEncoder(he))We$$

$$p(st \in V) = Softmax(BART(hm)Wd)$$

The model is trained to minimize the cross-entropy loss over the predicted tokens:

$$Lrec = -\sum_{t=1}^{t} \log p(s_t \in v)$$

All components are trained end-to-end. The CNN-Transformer pipeline functions as a surrogate encoder that learns to translate neural representations from EEG signals into a space compatible with textual embeddings, allowing the language model to decode coherent and semantically aligned sentences.

### 3.2 Zero-Shot Sentiment Classification from EEG

In this work, we propose an alternative framework for sentence-level sentiment classification using EEG signals, without relying on any explicit EEG-to-sentiment annotations. The approach operates in two stages: initially, a decoder converts EEG features into approximate textual representations;

subsequently, a sentiment classifier analyzes the generated text to infer the corresponding emotional label [1]. The core idea is to train the decoder and the sentiment classifier independently. The decoder is trained on EEG-to-text pairs, while the classifier is trained on external text-to-sentiment data [11]. This separation enables a zero-shot setup, as no direct EEG-to-sentiment supervision is required [17].

Figure 1.B shows the main architecture, by first decoding EEG into text, the system leverages the representational power of large language models for sentiment analysis without ever seeing labeled EEG–sentiment pairs. This modular strategy was first demonstrated in the open-vocabulary EEG-to-text and zero-shot sentiment work of Wang & Ji, who showed that intermediate text yields far better sentiment predictions than direct EEG classification [1].

## 4 Experiment

### 4.1 Dataset

In this work, we use the ZuCo dataset, a multimodal resource that captures brain activity and eye movement patterns during naturalistic reading scenarios [5], [7]. The dataset comprises synchronized EEG and eye-tracking signals collected from adult native English speakers. Participants engaged in two types of reading: general-purpose reading and a task-driven comprehension task. The texts span movie reviews and Wikipedia articles and include sentiment annotations for a subset of the data. To prepare the dataset for training, we applied preprocessing steps to extract both word-level and sentence-level neural features, aligning them with eye-tracking signals. Sentences containing invalid or missing values were discarded.

For model evaluation, the dataset was split at the sentence level to ensure that test data included entirely unseen content, with an (80%/10%/10%) split for training, validation, and testing.

| Reading Task | #Unique sentences | #Training Samples | #Test Samples |
|---|---|---|---|
| SR v 1.0 | 400 | 3391 | 418 |
| NR v 1.0 | 300 | 2406 | 321 |
| NR v 2.0 | 349 | 4456 | 601 |
| TSR v 1.0 | 407 | 3372 | 350 |

*Table 1: Dataset Statistics. SR: Normal Reading (sentiment), NR: Normal Reading (Wikipedia), TSR: Task Specific Reading (Wikipedia). We used data from 12 subjects in v1.0 and 18 subjects in v2.0.*

### 4.2 EEG-To-Text Decoding Evaluation

We trained the EEG-to-Text Sequence (ETS) model on the full ZuCo dataset suite—comprising both natural and task-specific reading conditions collected across 30 participants. The dataset includes SR (sentence reading), NR v1.0, NR v2.0 (natural reading), and TSR v1.0 (Task-Specific Reading), which provided the model with a rich variety of EEG and eye-tracking signals corresponding to diverse textual stimuli. This multimodal, multi-subject setup improves the generalization capabilities of the model and its robustness to individual variability in neural responses. To assess the quality of generated sentences, we adopted widely accepted text generation metrics: BLEU-N [8] and ROUGE-1 [9]. BLEU evaluates n-gram precision between generated and reference sentences, while ROUGE-1 captures unigram-level precision, recall, and F1, helping to evaluate content overlap.

### Results

Table 2 presents the primary results of our evaluation. The baseline approach by [1]. which utilizes a fine-tuned BART model, demonstrated moderate decoding performance, particularly struggling with higher-order n-gram matches. In contrast, ETS significantly advanced decoding performance, achieving BLEU-{1,2,3,4} scores of **43.39**, **30.36**, **23.67**, and **20.22**. These results correspond to improvements over the baseline of **+3.29** BLEU-1 (+8.2 %), **+7.26** BLEU-2 (+31.4 %), **+11.17** BLEU-3 (+89.4 %), and **+13.42** BLEU-4 (+197.4 %). The relative gain increases substantially for larger n-gram evaluations, indicating ETS's superior ability to model long-range syntactic and semantic coherence.

| Model name | BLEU-N(%) | | | | ROUGE-1(%) | | |
|---|---|---|---|---|---|---|---|
| | N=1 | N=2 | N=3 | N=4 | P | R | F |
| Baseline model(Wang & Ji) | 40.1 | 23.1 | 12.5 | 6.8 | 22.8 | 21.6 | 22.0 |
| ETS( Bart) | **43.39** | **30.36** | **23.67** | **20.22** | **37.70** | **35.77** | **36.66** |

*Table 2: presents the key results for the EEG-to-Text decoding task. When comparing our proposed model, ETS (Bart), against the baseline model.*

| Model name | BLEU-N(%) | | | | ROUGE-1(%) | | |
|---|---|---|---|---|---|---|---|
| | N=1 | N=2 | N=3 | N=4 | P | R | F |
| ETS(T5) | **48.15** | **31.93** | 22.15 | 16.29 | 37.54 | 35.27 | 36.27 |
| ETS (Bart) | 43.39 | 30.36 | **23.67** | **20.22** | **37.70** | **35.77** | **36.66** |

*Table 3: Performance comparison between our two models Bart and T5.*

Furthermore, ETS achieved ROUGE-1 scores of **37.7** (Precision), **35.77** (Recall), and **36.66** (F1), surpassing previous methods by **+6.00**, **+6.97**, and **+6.56** respectively. This further confirms that ETS more effectively captures essential content words and overall textual alignment.

These strong gains on higher-order metrics are particularly meaningful in the context of EEG-based decoding, where word-level neural signatures extracted at each eye fixation are often noisy or ambiguous due to re-reading, distraction, or other non-semantic cognitive processes [18].By aggregating context over longer spans and fusing gaze information, ETS smooths out these local inconsistencies, enabling more coherent multi-word predictions.

**T5 vs. BART Decoder**

In additional experiments, we replaced the BART decoder [3] with a T5 [19] architecture to explore the impact of different pretrained language models. Interestingly, T5 outperformed BART on BLEU-1 (+4.76; +10%) and BLEU-2 (+1.57; +5%). This suggests that T5 may offer better token-level alignment in the early stages of sentence construction, likely due to its encoder-decoder architecture optimized for transfer tasks. However, BART remains competitive, particularly for longer and more structured sentence generation, as shown in Table 3.

**4.3 EEG-Based and Zero-Shot Sentiment Classification**

**Experimental Setup**

We evaluate two approaches for sentiment classification using EEG data from the Reading Task SR v1.0 dataset. Each instance in the dataset is a triplet $\langle E,S,c \rangle$, where $E$ represents EEG signals recorded while reading, $S$ is the corresponding sentence, and $c$ is the associated sentiment label.

**Approach 1: Direct EEG-to-Sentiment Classification**

The first approach directly predicts sentiment labels from EEG signals without relying on textual representations. To do so, we reuse the EEG Encoder presented in Table 2.A, making a single key modification: we replace the text-generation head of the decoder with

| 1 | Ground Truth: He **was** United **States Secretary of Defense** from 1987 until 1989.<br><br>Model Output: **was** a **States Secretary of Defense** in 1981 to 1993. |
|---|---|
| 2 | Ground Truth: He **was** <u>inducted</u> **into the** <u>College</u> Football **Hall of Fame**, the South Carolina <u>Sports</u> **Hall of Fame**, **and the** Clemson Ring of Honor.<br><br>Model Output: **was** <u>educateded</u> **into the** United of **Hall of Fame** in **the** highest **Carolina** <u>Athletic</u> **Hall of Fame, and the** Georgia <u>University</u> of Fame. |
| 3 | Ground Truth: The arm injury kept **him from** being drafted into the <u>armed</u> **forces** for **the** <u>Second</u> **World War.**<br><br>Model Output: movie of and **him from** playing able by the <u>Army</u> **forces**, **the** <u>duration</u> **World War**. |
| 4 | Ground Truth: He graduated in <u>1937</u> **from** the **University of** California, **Berkeley with a** <u>degree</u> **in economics** and philosophy, and earned a master's **degree** from the Harvard Graduate School of <u>Business</u> Administration **in** <u>1939</u>.<br><br>Model Output: was **from** <u>1983</u> with New **University of** Wisconsin, **Berkeley with a** <u>Bachelor</u> **in economics**, was. and was a scholarship's **degree** in the University University **School of** <u>Management</u> **in** in <u>1938</u>. |

*Table 4: EEG-To-Text decoding examples on unseen test data*

This allows the model to learn a direct mapping from EEG signals $E$ to sentiment labels $c$ while preserving the structural backbone of the original EEG encoder.

The model is trained end-to-end using $\langle E,c \rangle$ pairs [10], [15]. Despite fitting EEG-sentiment data directly, this approach is constrained by noisy mappings and limited labeled data.

**Approach 2: Zero-Shot EEG-to-Sentence-to-Sentiment**

In contrast, the second approach follows a zero-shot pipeline. EEG signals are first decoded into text using the pretrained EEG-to-Text model. The resulting sentence predictions are then passed to a sentiment classifier trained on external text datasets. This method avoids the need for any $\langle E,c \rangle$ training pairs and benefits from powerful, pretrained text-based sentiment models.

We evaluate multiple variants of this pipeline using different decoder and classifier combinations.

**Text-Only Baselines**

For reference, we also evaluate sentiment classifiers trained and tested on textual inputs $\langle S,c \rangle$ using large language models like BERT [17], RoBERTa [20], and BART [3]. These baselines provide an upper bound on performance by directly operating on human-readable text

**Evaluation Protocol**

All models are evaluated on the same held-out test set from the SR v1.0 dataset. This consistent evaluation enables a fair comparison between EEG-based, zero-shot, and text-based approaches.

**Results and Observations**

Table 5 show that the direct EEG-to-Sentiment classifier achieves reasonable performance but is limited by the noisiness and low signal-to-label mapping of brain data. On the other hand, the zero-shot EEG-to-Text-to-Sentiment pipeline substantially outperforms the direct method, benefiting from intermediate textual representations and the strength of pretrained language model.

| Model | Test on | Precision(%) | Recall(%) | F1(%) | Accuracy(%) |
|---|---|---|---|---|---|
| EEG Encoder+Bert | C | 37.48 | 37.08 | 35.86 | 38.31 |
| EEG Encoder+Roberta | C | 40.38 | 38.98 | 37.14 | 40.56 |
| ETS+Roberta | PS | 37.27 | 42.35 | 36.30 | 39.02 |
| ETS+Bert | PS | 41.66 | 39.38 | 38.95 | 40.02 |
| ETS+Bart | PS | **70.04** | **69.55** | **68.18** | **68.23** |
| Roberta | RS | 64.10 | 65.53 | 56.39 | 61.43 |
| Bert | RS | 68.52 | 70.00 | 67.98 | 68.34 |
| Bart | RS | **95.72** | **96.27** | **95.83** | **95.85** |

*Table 5* Ternary sentiment classification results on the SR v1.0 test set, comparing the direct EEG-to-sentiment approach with the application of a pre-trained sentiment analysis model on both the predicted and ground-truth sentences. In test input, C : is the unseen data from SR v1.0, PS: is the predicted sentences from ETS on SR v1 unseen data, Rs: is the real sentences from SR v1 unseen data.

## 5. Related Work

### 5.1 EEG-Based Brain-to-Text Decoding

Translating brain activity into text has been a longstanding challenge in BCI research. Early studies primarily relied on classification techniques for restricted vocabularies, using handcrafted features from EEG signals [21], [22]. More recent approaches shifted toward open-vocabulary generation by leveraging neural language models. Wang and Ji [1] introduced a fine-tuned BART model for EEG-to-text generation, achieving moderate BLEU scores but struggling with higher n-gram precision. Duan et [2] proposed DeWave, a discretized EEG encoding scheme, which showed performance gains using a codebook-style representation. However, these approaches still face challenges in modeling long-range dependencies and capturing semantic coherence, which our ETS model aims to overcome.

### 5.2 Multimodal Decoding with Eye-Tracking

Combining EEG with eye-tracking signals has proven effective in improving decoding accuracy. The ZuCo dataset [5] pioneered this multimodal setup, providing synchronized EEG and gaze data during naturalistic reading tasks. Eye movements provide crucial temporal and semantic context, enhancing the alignment between neural signals and language. Models like [23] utilized these signals for sentiment classification and reading task detection. Our work extends this by jointly modeling EEG and gaze features for sentence-level text generation, demonstrating improved generalization and robustness to inter-subject variability.

### 5.3 Sequence-to-Sequence Models for Cognitive Decoding

Transformer-based models have revolutionized NLP, and their application to EEG-to-text decoding is gaining attention.

Traditional Seq2Seq frameworks using RNNs or LSTMs were limited by vanishing gradients and short-term memory. The advent of Transformer architectures [6] enabled better modeling of long-range dependencies. In our work, we adopt a Transformer-based encoder for EEG encoding and explore both BART [3] and T5 [19] decoders for text generation. We show that such architectures outperform prior generative models, especially for higher-order n-gram coherence.

### 5.4 Sentiment Classification from EEG

Sentiment decoding from EEG has typically been approached via classification rather than generation. Prior work focused on emotional state detection using frequency-domain features and shallow classifiers [24].Deep learning models like EEGNet [15] extended these efforts to end-to-end pipelines. However, few works explore sentence-level sentiment classification using multimodal signals. Our ETS framework fills this gap by integrating sentiment recognition within a generative EEG-to-Text architecture, offering an end-to-end solution that bridges affective and linguistic decoding.

### 5.5 Inner Speech and Thought Decoding

Inner speech decoding translating imagined or subvocalized speech into text shares conceptual overlap with EEG-to-text. Yet, datasets for inner speech [21], [25] remain small, domain-restricted, and often lack sentence-level annotations. While invasive methods (e.g., intracranial recordings) show promising results [12], non-invasive approaches are limited by signal noise and vocabulary constraints. Our work offers a foundational step toward open-vocabulary, non-invasive inner speech decoding by demonstrating sentence-level generation from EEG, laying the groundwork for more advanced inner speech applications.

## 6 Conclusion

In this paper, we introduced a novel **open-vocabulary EEG-to-Text Sequence-to-Sequence (Seq2Seq) decoding framework**, alongside an EEG-based **sentence-level sentiment classification** model. Our proposed architecture, **ETS** demonstrated significant performance improvements over existing approaches in both neural language decoding sentiment classification.

Despite these advancements, practical deployment of such systems in real-world applications remains limited by factors such as **dataset size, noise in neural signals**, and **participant variability**. The scarcity of large-scale, high-quality neural datasets continues to hinder progress, especially in open-domain text generation from EEG.

A promising direction for future research is to adapt the ETS framework to the task of **inner speech decoding** in an open-vocabulary context. Current inner speech datasets, such as [24] offer limited lexical diversity and sentence-level coverage. To enable robust decoding of imagined or subvocalized thoughts, the development of **larger, sentence-aligned inner speech datasets with broader vocabulary** will be essential. Expanding the model to such challenging domains could further bridge the gap between brain activity and language, opening avenues for non-invasive neural communication technologies.

## Acknowledgement

We would like to thank our professor, **Dr. Hala Abdel Galil**, for her invaluable guidance, support, and feedback throughout this project. Her expertise and encouragement were instrumental in shaping the direction and quality of this work.